\documentclass[11pt,a4paper]{article}
\usepackage[hyperref]{emnlp2020}
\usepackage{times}
\usepackage{latexsym}

\usepackage{multirow}
\usepackage{amsmath}
\usepackage{amssymb}
\usepackage{algorithm}
\usepackage{algorithmic}

\usepackage{soul}
\usepackage{color}

\usepackage{microtype}
\usepackage{xcolor}
\usepackage{amsmath}
\usepackage{graphicx}
\definecolor{trolleygrey}{rgb}{0.5, 0.5, 0.5}

\title{Asking Complex Questions with Multi-hop Answer-focused Reasoning}

\author{Xiyao Ma$^1$, Qile Zhu$^1$, Yanlin Zhou$^1$, Xiaolin Li$^2$, Dapeng Wu$^1$ \\
  $^1$NSF Center for Big Learning, University of Florida \\
  $^2$Cognization Lab \\
  { maxiy@ufl.edu, valder@ufl.edu, zhou.y@ufl.edu, xiaolinli@ieee.org, dpwu@ufl.edu}}

\date{}

\begin{document}
\maketitle
\begin{abstract}
Asking questions from natural language text has attracted increasing attention recently, and several schemes have been proposed with promising results by asking the right question words and copy relevant words from the input to the question.
However, most state-of-the-art methods focus on asking simple questions involving single-hop relations. 
In this paper, we propose a new task called \textbf{multi-hop question generation} that asks complex and semantically relevant questions by additionally discovering and modeling the multiple entities and their semantic relations given a collection of documents and the corresponding answer \footnote{Our work is similar to \citet{pan2020semantic}, and we propose the similar noval challenging task on HOTPOTQA dataset independently. The major differences are: 1. We built the graph for reasoning following different heuristics. Specifically,  \citet{pan2020semantic} maily adopted SRL and dependence parse tree, while we utilized NER, coreference resolution, and surface matching. 2. \citet{pan2020semantic} included all the data examples in HOTPOTQA dataset for training and validation. However, after deep diving into the dataset, we argued that the questions with type of "comparison" are not suitable for the proposed task as they do not require QG models to discover and gather mutli-hop semantic relations among the entities. 3. Different from \cite{pan2020semantic} where no testing dataset is 
available to evaluate models, we proposed to combine the training and dev dataset together and split them into traininng, dev, and testing dataset. Please find the detail in the Section \ref{dataset}. The dataset and code are available at \url{https://github.com/Shawn617/Multi-hop-NQG}}.
To solve the problem, we propose multi-hop answer-focused reasoning on the grounded answer-centric entity graph to include different granularity levels of semantic information including the word-level and document-level semantics of the entities and their semantic relations.
Through extensive experiments on the HOTPOTQA dataset, we demonstrate the superiority and effectiveness of our proposed model that serves as a baseline to motivate future work.
\end{abstract}

\section{Introduction}
Given a background context and the corresponding answer, the question generation (QG) task aims to ask a semantically relevant question. 
QG has considerable benefits in education scenario, dialogue systems, and question answering \cite{du2017learning}.
Recently, many approaches have been proposed to solve the problem \cite{zhou2017neural, sun2018answer, ma2019improving}, mostly realized by variants of the seq-to-seq model \cite{sutskever2014sequence} with attention and copy mechanism \cite{cho2014learning, bahdanau2014neural}.

However, existing works mainly focus on asking a simple question $Y=\{y_t\}^N_{t=1}$ by only capturing one direct relation among the entities from the context input $X=\{x_t\}^M_{t=1}$.
Taking one example from SQuAD dataset \cite{rajpurkar2016squad} as shown in the upper part of the Table \ref{tab:task}, the model only needs to capture the single-hop relation between entity ``Donald Davies" and the answer entity ``Message Routing Methodology" and ask the question ``What did Donald Davies develop?"

\begin{table}[ht]
\centering
\caption{Comparison of the single-hop question generation task on the SQuAD dataset \cite{rajpurkar2016squad} and the proposed multi-hop question generation task on the HOTPOTQA dataset \cite{yang2018hotpotqa}}.
\scalebox{0.88}{
\begin{tabular}{l}
    \hline
    \textbf{Single-hop Question Generation}\\
    \textbf{Document}: starting in 1965, Donald Davies at the \\ National Physical Laboratory, UK, independently \\
    developed the same \textcolor{red}{Message Routing Methodology}\\
    as developed by baran.\\
    \textbf{Question}: \textit{What did Donald Davies develop?}\\
    \hline
    \textbf{Multi-hop Question Generation}\\
    \textbf{Document 1}: [Peggy Seeger] Margaret ``Peggy" \\
    Seeger (born June 17, 1935) is an \textcolor{red}{American} folksinger. \\ 
    She is also well known in Britain, where she has lived \\
    for more than 30 years, and was married to the singer \\
    and songwriter Ewan MacColl until his death in 1989.\\
    \textbf{Document 2}: [Ewan MacColl] James Henry Miller \\
    (25 January 1915 – 22 October 1989), better known by \\
    his stage name Ewan MacColl, was an English folk \\
    singer, songwriter, communist, labour activist, actor, \\
    poet, playwright and record producer.\\
    \textbf{Question}: \textit{What nationality was James Henry}\\
    \textit{Miller's wife?}\\
    \hline
\end{tabular}}
\label{tab:task}
\end{table}

In this paper, we propose a new task called multi-hop neural question generation.
Given a collection of documents $\mathcal{D}=\{d_i\}^I_{i=1}=\{X_{text}^i, X_{title}^i\}^I_{i=1}$ each containing a context $X_{text}^i$ and a title $X_{title}^i$, assuming that the answer $A$ exists in at least one document, the model aims to generate a complex and semantically relevant question $Y=\{y_t\}^N_{t=1}$ with multiple entities and their semantic relations.
One example is shown in the lower part of Table \ref{tab:task}.
The model need to discover and capture the entities (e.g., ``Peggy Seege", ``Ewan MacColl", and ``James Henry Miller") and their relations (e.g., ``Peggy Seege" marrited to ``James Henry Miller", and ``James Henry Miller" is the stage name is ``Ewan MacColl"), then ask the question ``What nationality was James HenryMiller’s wife?" according to the answer ``American".

In addition to the common challenges in the single-hop question generation task that the model needs to understand, paraphrase, and re-organize semantic information from the answer and the background context, another key challenge is in discovering and modeling the entities and the multi-hop semantic relations across documents to understand the semantic relation between the answer and background context.
Merely applying a seq-to-seq model on the document text does not deliver comparable results in that the model performs poorly on capturing the structured relations among the entities through multi-hop reasoning.

In this paper, We propose the multi-hop answer-focused reasoning model to tackle the problem.
Specifically, instead of utilizing the unstructured text as the only input, we build an answer-centric entity graph with the extracted different types of semantic relations among the entities across the documents to enable the multi-hop reasoning.
Inspired by the success of graph convolutional network (GCN) models, we further leverage the relational graph convolutional network (RGCN) \cite{schlichtkrull2018modeling} to perform the answer-aware multi-hop reasoning by aggregating the different levels of answer-aware contextual entity representation and semantic relations among the entities.
The extensive experiments demonstrate that our proposed model outperforms the baselines in terms of various metrics.
Our contributions are three-fold:
\begin{itemize}
    \item To the best of our knowledge, we are the first to propose the multi-hop neural question generation task,  asking complex questions from a collection of documents through multi-hop reasoning.
    \item We propose a multi-hop answer-focused reasoning model to dynamically reason and aggregate different granularity levels of answer-aware contextual entity representation and semantic relations among the entities in the grounded answer-centric entity graph.
    \item We conduct extensive experiments to demonstrate that our proposed model outperforms SOTA single-hop QG models and graph-based multi-hop QG model in terms of the main metrics, downstream multi-hop reading comprehension metrics, and human judgments.
    Our work offers a new baseline and motivates future researches on the task.
\end{itemize}

\section{Methods} 
\begin{figure*} [t]
    \centering
    \includegraphics[width=0.9\textwidth]{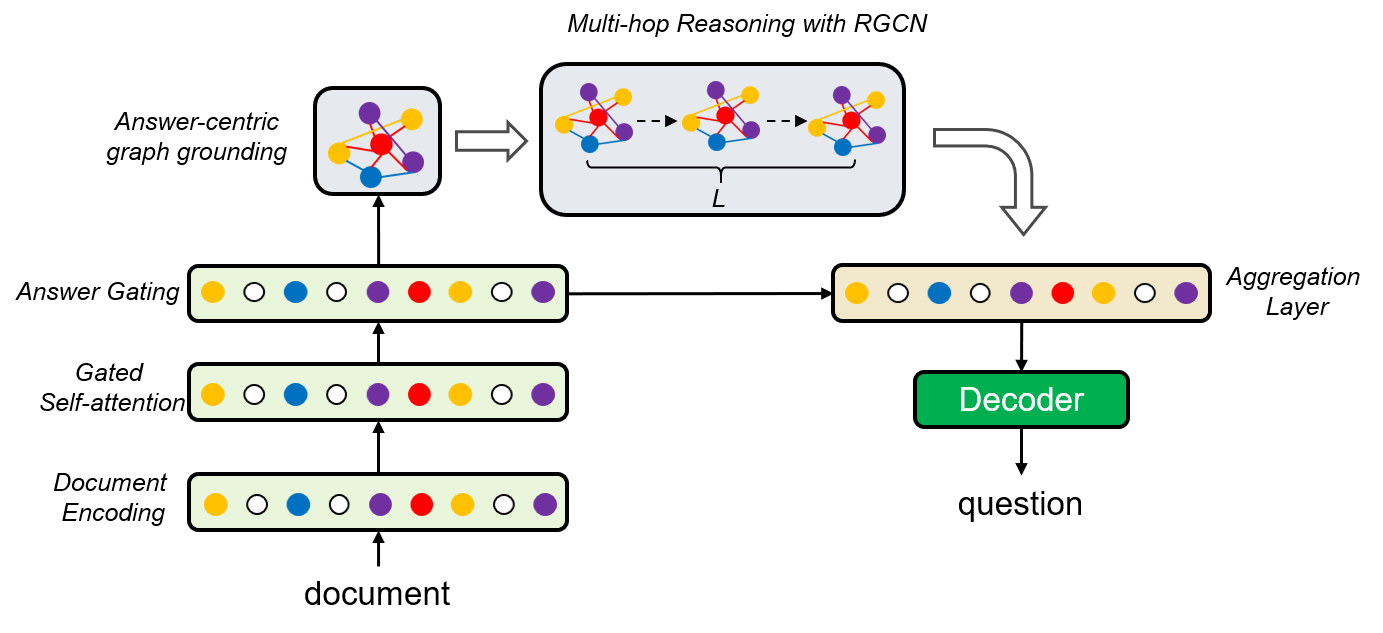}
    \caption{Architecture of Answer-focused Multi-hop Reasoning Model.}
    \label{fig:model}
\end{figure*}

In this section, we present the architecture and each module of the proposed multi-hop answer-focused reasoning model. 
The overall architecture is shown in Figure \ref{fig:model}. 
Our proposed method adopts a seq-to-seq  backbone \cite{sutskever2014sequence} incorporated with attention and copy mechanism \cite{bahdanau2014neural, gulcehre2016pointing}. 
Our model can be categorized into three parts: (i) answer-focused document encoding, (ii) multi-hop answer-centric reasoning, and (iii) aggregation layer, finally providing an answer-focused and enriched contextual representation.

\subsection{Answer-focused Document Encoding}
\paragraph{Document Encoding} Given input documents to the model, we represent them as a sequence of words $\mathbf{X}=\{x_i\}^M_{i=1}$ by concatenating the text words $X_{text}^i$ and the title words $X_{title}^i$ of each document:
\begin{align}
    \mathbf{X}=\left\{X_{text}^0, X_{title}^0, ... , X_{text}^I, X_{title}^I\right\},
\end{align}

Following \cite{zhou2017neural}, for each word $x$, we obtain its embedding by concatenating its word embedding, answer positional embedding, and feature-enriched embedding (e.g., POS, NER).

An one-layer bi-directional LSTM \cite{hochreiter1997long} is utilized as the encoder to obtain the document representation $H=[h_1,h_2,...,h_m] \in \mathbb{R}^{M*D}$:
\begin{align} 
    & h_i = \operatorname{LSTM}_{enc}(x_i, {h_{i-1}}).
\end{align}

\paragraph{Gated Self-attention Layer} The above document representation has limited knowledge of the context \cite{wang2017gated}. 
The gated self-attention layer is utilized to learn a contextual document representation $\hat{h}_{i}$ with a Bi-GRU \cite{chung2014empirical}:

\begin{align}
    \hat{h}_{i}=\operatorname{Bi-GRU}\left(\hat{h}_{i-1}^D,\left[h_{i}, o_{i}\right]\right),
\end{align}
where $v_i$ is the contextual vector obtained by attending to the context:
\begin{align}
d_{j}^{i} &= \mathrm{W_d}^{\mathrm{T}} \tanh (W'_{v} h_{j}+W_{v} h_{i}), \\
a_{k}^{i} &= \exp (d_{k}^{i}) / \Sigma_{j=1}^{n} \exp (d_{j}^{i}), \\
o_{i} &= \Sigma_{k=1}^{n} a_{k}^{i} h_{k}.
\end{align}

where $W_d$, $W_v$, and $W'_v$ are the trainable weights in the neural networks.

\paragraph{Answer Gating Mechanism} We further propose the answer gating mechanism to empower the model to learn the answer-focused document representation $H^a=\{\hat{h}^a_i\}^M_{i=1}$.
Utilizing a gate computed by a $sigmoid$ function to control the information passing, only the answer-related semantic information of the documents is forwarded for the downstream multi-hop reasoning:
\begin{align} 
    & {h}_i^a = \sigma(a W_a \hat{h}_i) * \hat{h}_i.
\end{align}
where the answer vector $a\in \mathbb{R}^{D}$ is the hidden state of the first answer word, and $W_a$ is a trainable parameters in the bilinear function.

\subsection{Multi-hop Answer-focused Reasoning}

\paragraph{Answer-centric Entity Graph Grounding} \label{sec:graph_building}

\begin{figure} [t]
    \centering
    \includegraphics[width=0.5\textwidth]{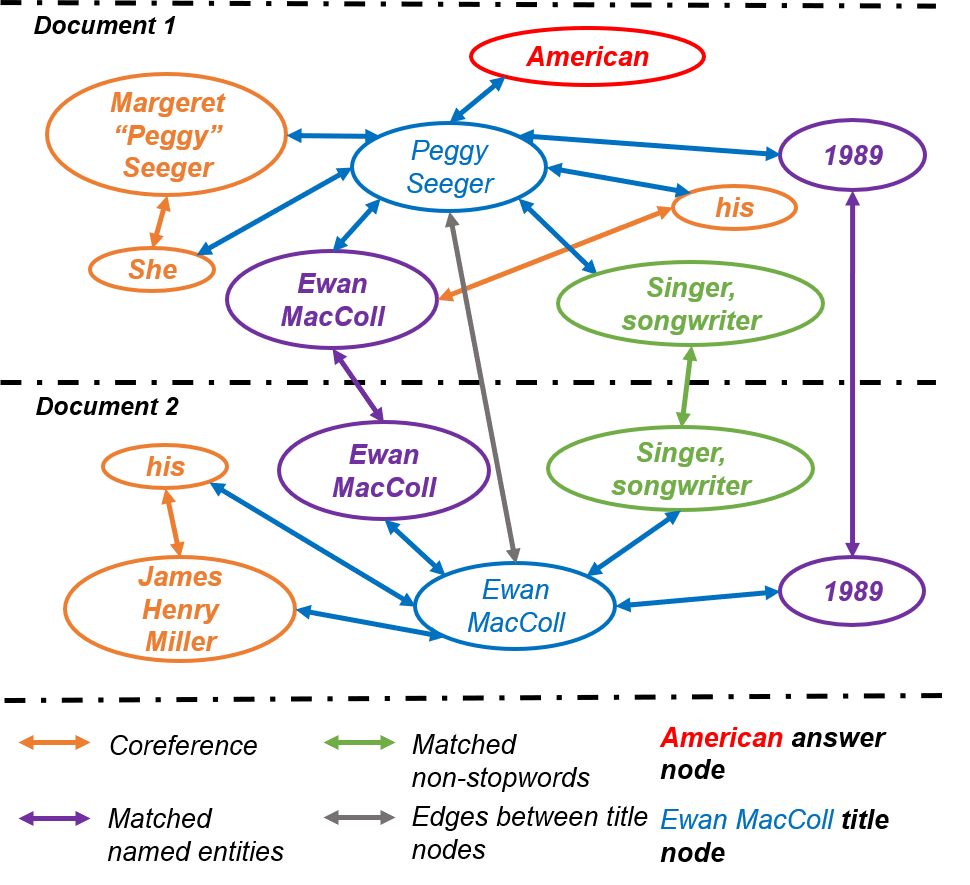}
    \caption{Diagram of an answer-centric entity graph example $\mathcal{G}=\{\mathcal{V}, \mathcal{E}\}$ built on the documents in Table \ref{tab:task}.
    The text in ovals and the solid lines in different colors indicate the different semantic types of the nodes $\mathcal{V}$ and the edges $\mathcal{E}$, respectively.
    The answer node is connected with all other nodes in the graph, which are not drawn for conciseness.
    }
    \label{fig:entity_graph}
\end{figure}

To explicitly discover and model the multiple entities and their semantic relations across documents, we ground an answer-centric entity graph from the unstructured text.

Let an answer-centric entity graph be denoted as $\mathcal{G}=\{\mathcal{V}, \mathcal{E}\}$, where $\mathcal{V}$ denotes entity nodes at different levels, and $\mathcal{E}$ denotes the edges between nodes annotated with different semantic relations.

To do so, we first exploit the Spacy toolkit \cite{honnibal2017spacy} to extract the named entity and all coreference words.
Then, we identify the exactly matched non-stop words from the documents.
We treat these exactly matched non-stop words, named entities, the answer and titles as the nodes in the answer-centric entity graph, which represent different granularity levels of the contextual representation: (1) the exactly matched non-stop words and entity nodes encode the word-level and local representation in the specific document context; (2) the title nodes represent the document-level semantics; (3) the answer node offers the answer-aware representation for the graph reasoning, and models a global representation across documents.

We then define edges between two nodes by leveraging different types of semantics within the documents in the following heuristic: 

\vspace{0.1cm}

(1) We connect all exactly matched named entities no matter they are in the same documents or different documents (e.g., ``Ewan MacColl").

\vspace{0.1cm}

(2) We connect all inter-document and intra-document exact matched non-stop words (e.g. ``singer, songwriter"). 

\vspace{0.1cm}

(3) All coreference words are then linked to each other.

\vspace{0.1cm}

(4) We further connect the title node with all entity nodes within the same document.

\vspace{0.1cm}

(5) We add dense connections between all title nodes.

\vspace{0.1cm}

(6) The answer node is connected to all other nodes in the graph, resulting in an answer-centric entity graph.

\vspace{0.1cm}

An example graph built from the documents of the example in Table \ref{tab:task} is shown in Figure \ref{fig:entity_graph}, representing different granularity levels of the semantic information with various nodes and edges.


\paragraph{Multi-hop Reasoning with RGCN}
To make use of the grounded answer-centric entity graph, we leverage GNN-based model to conduct the multi-hop reasoning.
In general, with different message passing strategies, graph neural network-based models update the node representation based on its first-order neighbors.

Specifically, we employ the RGCN for the multi-hop reasoning \cite{schlichtkrull2018modeling}.
We first initialize the representation of node $v \in \mathcal{V}$ with the output from answer gating mechanism by $v^0_i = h^a_j$ or $v^0_i = average(h^a_j, h^a_j+1, ..., h^a_k)$ if the entity node contains multiple words.
Meanwhile, edges are annotated by one-hot vectors indicating different semantic relations.
In each layer $0 \leq \ell \leq L$, the representation of node $i$ is updated by the summation of the transformation of its node representation and the transformation of its neighbors:
a\begin{align}
    v_{i}^{(l+1)}=\sigma\left(\sum_{r \in \mathcal{R}} \sum_{j \in \mathcal{N}_{i}^{r}} \frac{1}{c_{i, r}} W_{r}^{(l)} v_{j}^{(l)}+W_{0}^{(l)} v_{i}^{(l)}\right),
\end{align}
where $W_{r}^{(l)}$ is relation-specific trainable weights. 
The number of parameters of the weights is further decreased by the linear combination of a basis weight $W_{b}^{(l)} \in \mathbb{R}^{d^{(l+1)} \times d^{(l)}}$ and relation-specific coefficients $a_{r b}^{(l)}$:
\begin{align}
    W_{r}^{(l)}=\sum_{b=1}^{B} a_{r b}^{(l)} W_{b}^{(l)}.
\end{align}
After $L$ layers of reasoning, at most $L$-hop relations can be captured.

\subsection{Aggregation Layer}
Inspired by \cite{peters2018deep}, the final answer-aware contextual representation is computed by selectively aggregating the output of each RGCN layer and the answer-aware document representation generation with trainable layer-wise weights.
Similarly, the answer node representation of each layer and the last hidden state of the LSTM are stacked together to produce a more accurate document-level and global representation:
\begin{align}
    & H_{\mathcal{G}} = W_c ([v^1, v^2, ..., v^L, H^a]), \\
    & z = W_g ([v_a^1, v_a^2, ..., v_a^L, h^a_M]).
\end{align}
where $V_\ell = [v_1, v_2, ... , v_M]$ is node representations of the $\ell$th layer, and $\hat{v}_a^{\ell}$ is the answer node representation of the $\ell$th layer.
The $W_c$ and $W_g$ are the layer-wise trainable weights.
By doing so, the different granularities of contextual representations expressing various types of semantics are aggregated to produce the final entity-level $H_{\mathcal{G}} \in \mathbb{R}^{M*D}$ and document-level $z \in \mathbb{R}^D$ representation for the decoder.

\subsection{Decoder}
With a hidden state initialized to $s_0=z$, a uni-directional LSTM is utilized as the decoder to generate the question, where the current hidden state is updated given the previous generated word and the previous hidden state:
\begin{align} 
    & s_t = LSTM_{Dec}([w_t; c_{t-1}], s_{t-1}),
\end{align}
where the context vector $c$ is computed with the attention mechanism \cite{bahdanau2014neural} by attending to the encoder hidden states:
\begin{align}
    & e_t = H_{\mathcal{G}}^T W_e s_t, \\
    & \alpha_t = Softmax(e_t), \label{softmax} \\
    & c_t = H_{\mathcal{G}}^T \alpha_t.
\end{align}

To solve the Out-of-Vocabulary issue, we also exploit the copy mechanism to steer the model to copy a word from the input \cite{see2017get, gulcehre2016pointing}.
Specifically, in each step in the decoder, a probability is computed, deciding to copy words from the input documents based on attention matrix or generate a word from the vocabulary via an output layer with softmax function:
\begin{align}
 & g_{copy}=\sigma(W^cs_t + U^cc_t + b^c), \\
 & p_{generate}(y_t) = Softmax(f(s_t, c_t)).
\end{align}

Finally, treating the copy probability as the attention weights (e.g., $P_{copy} = \alpha_t$), the final word distribution is the summation of the probability of generating a word from the vocabulary and the probability of copying a word from input:
\begin{equation}
\begin{split}
    p_{\text {final}}&\left(y_{t} | y_{<t}; \theta_{s2s}\right)=g_{copy} p_{\text {copy}}(y_t, )\\
    &+\left(1-g_{\text {copy}}\right) p_{\text {generate}}(y_t).
\end{split}
\end{equation}


\section{Experiment}
In this section, we conduct extensive experiments on the HOTPOTQA dataset \cite{yang2018hotpotqa}, demonstrating the performance of the proposed model by comparing it with the existing SOTA single-hop question generation models and a multi-hop question generation model with GAT \cite{velivckovic2017graph}.

\subsection{Experiment Setting}

\paragraph{Dataset} \label{dataset}
HOTPOTQA dataset is an accessible dataset collected from Wikipedia articles for the multi-hop reading comprehension task \cite{yang2018hotpotqa}. 
We discard the questions with the ``comparison" type, and we only collect the text labeled ``supporting facts'' in the set of documents.
Lack of access to the original testing dataset, we combine the training set and development set and randomly split them into the training set, development set, and testing set with the size of 68758, 4992, and 4991 data samples, respectively.

\begin{table*}[t]
\centering
\caption{Comparison of models performances in terms of the main metrics on HOTPOTQA dataset.}
\scalebox{1}{
\begin{tabular}{l|cccccc}
\hline
Models & BLEU-1 & BLEU-2 & BLEU-3 & BLEU-4 & METEOR & ROUGE-L  \\ 
\hline
NQG++ \cite{zhou2017neural} & 44.55& 33.18 & 26.57 & 21.99 & 24.35 & 41.08 \\ 
PG \cite{see2017get} & 46.13 & 35.14 & 28.71 & 24.12 & 24.14 & 42.18  \\
SM-API \cite{ma2019improving} & 46.95 & 35.76 & 29.02 & 24.34 & 24.30 & 42.32  \\
\hline
PG + GAT & 47.35 & 36.10 & 29.85 & 24.98 & 24.56 & 42.62 \\
\textbf{Proposed} & \textbf{50.93} & \textbf{38.93} & \textbf{31.78} & \textbf{26.70} & \textbf{25.40} & \textbf{43.88} \\
\hline
\end{tabular}}
\label{tab:mainMetrics}
\end{table*}

\paragraph{Baselines} In the experiments, we compare the performance of our proposed model and several baseline models as follows:
\begin{itemize}
    \item NQG++ \cite{zhou2017neural}: It is a commonly used baseline for the single-hop neural question generation task.
    The concatenated document text is passed into the seq-to-seq model with the answer positional embedding and enriched lexical features (e.g., named entity, pos-of-tag, and case).
    Attention and copy mechanisms are adopted in the decoder.
    \item Pointer-generator (PG) \cite{see2017get}: Originally proposed for text summarization task, it is revised to solve the question generation problem. 
    The copy mechanism is realized differently.
    We also add enriched lexical features in the embedding layer like NQG++. 
    \item Sentence-level Semantic Matching and Answer Position Inferring (SM-API) \cite{ma2019improving}: It is a state-of-the-art model on the single-hop neural question generation task.
    It proposes two modules called sentence-level semantic matching and answer position inferring, trained jointly with the seq-to-seq model to ask questions containing the right question words, keywords, and answer-aware semantics.
    \item PG + GAT: Graph attention network \cite{velivckovic2017graph} updates the node representation by attending to the representation of its neighbors.
    One straight forward way for multi-hop reasoning is to utilize three layers of the graph attention model (GAT) on the built answer-centric entity graph illustrated in Section \ref{sec:graph_building}.
\end{itemize}

\subsection{Results and Analysis}

\paragraph{Main Metrics} We evaluate model performances in terms of BLEU1-4 \cite{papineni2002bleu}, METEOR \cite{denkowski2014meteor}, and ROUGE-L \cite{lin2004rouge} on HOTPOTQA dataset in Table \ref{tab:mainMetrics}.

SM-API model only improves the PG model by $0.22$ on the BLEU-4 score and does not show considerable advantage on the multi-hop question generation task.
This is in part due to that the answer position inferring module designed explicitly for the single-hop answer position prediction does not offer a very accurate supervised signal for the model training on the multi-hop dataset.
On the other hand, the dataset does not include data samples where questions about different answers are asked given the same context, thus limits the power of the sentence-level semantic matching module.

Stacking several layers of GAT directly on the LSTM encoder improves the performance by leveraging the answer-centric entity graph for multi-hop reasoning; nevertheless the different semantic relations and answer-focused entity representations are ignored during the multi-hop reasoning.

Our proposed multi-hop answer-focused reasoning model achieves much higher scores than the baselines as it leverages different granularity levels of answer-aware contextual entity representation and semantic relations among the entities in the grounded answer-centric entity graph, producing precise and enriched semantics for the decoder.

\paragraph{Downstream Task Metrics} Main metrics has some limitations as they only prove the proposed model can generate questions similar to the reference ones.
We further evaluate the generated questions in the downstream multi-hop machine comprehension task.

Specifically, we choose a well-trained DecompRC \cite{min2019multi}, a state-of-the-art model for multi-hop machine comprehension problem on the same HOTPOTQA dataset, to conduct the experiment.
In general, DecompRC decomposes the complex questions requiring multi-hop reasoning into a series of simple questions, which can be answered with single-hop reasoning.
The performance of DecompRC on different generated questions reflect the quality of generated questions and the multi-hop reasoning ability of the models, intuitively.

\begin{table}[ht]
\centering
\caption{Performance of the DecompRC model in the downstream machine comprehension task in terms of EM and F1 score.}
\scalebox{0.9}{\begin{tabular}{l|lr}
\hline
Questions     & EM (\%) & F1 (\%) \\ 
\hline
Reference Questions        & 71.84   &  83.73  \\
\hline
NQG++ \cite{zhou2017neural}  &  65.82  &  76.97  \\
PG \cite{see2017get}  & 66.70  &  78.03  \\
SM-API \cite{ma2019improving}  & 67.01  &   78.43  \\
\hline
PG + GAT   &  67.23  &  79.01  \\
\textbf{Proposed} &  \textbf{69.92} &  \textbf{81.25}\\
\hline
\end{tabular}}
\label{tab:MCmetrics}
\end{table}

We report the Exact Match (EM) and F1 scores achieved by the DecompRC model in Table \ref{tab:MCmetrics}, given the reference questions and different model-generated questions. 
The human-generated reference questions have the best performance.
The DecompRC model achieves a much higher EM and F1 scores on the questions generated by our proposed model than the baseline models.

\paragraph{Analysis of Answer-focused Multi-hop Reasoning}
The design of the answer-focused multi-hop reasoning model is to discover and capture the entities relevant to the answer utilizing the various types of the semantic relations among them.
We analyze the model effect by measuring the quantity of named entities in the generated question in terms of the Precision and Recall, similar to \cite{sun2018answer}.
Quantitatively, Given a generated question $G=\{g_i\}^N_{i=1}$ and its reference question $R=\{r_i\}^N_{i=1}$, we define:
\begin{align}
    \text{Precision} =& \frac{\text{\# NEs in both $g_i$ and $r_i$}}{ \text{\# NEs in $g_i$}}  \\
    \text{Recall} =& \frac{ \text{\# NEs in both $g_i$ and $r_i$}}{ \text{\# NEs in $r_i$}}
\end{align}
where $\#\text{NEs}$ indicates the number of named entities.

\begin{table}[ht]
\caption{Comparison of Precision and Recall on different model-generated questions.}
\centering
\scalebox{0.9}{
\begin{tabular}{l|cc}
\hline
Models &  Precision & Recall \\
\hline
NQG++   &  46.59 &  52.66 \\
PG  & 46.64 &  52.45 \\
SM-API & 46.82 &  53.10 \\
 \hline
PG + GAT & 47.30 &  53.44 \\
\textbf{Proposed} &  \textbf{49.29} &  \textbf{54.64} \\
\hline
\end{tabular}}
\label{tab:ner_eval}
\end{table}

As reported in Table \ref{tab:ner_eval}, our proposed model outperforms the baselines, indicating that our model can generate questions involving more answer-aware entities by leveraging the answer-focused multi-hop reasoning.


\subsection{Human Evaluation}
We further examine 100 generated questions with human evaluation by scoring them on a scale from 1 to 5 in the light of semantic relatedness, fluency, and complexity.
Semantic relatedness measures how well a generated question matches with the documents and the answer.
Fluency reflects the naturalness of the generated questions, and complexity measures whether the generated questions are complicated and involve multiple entities.

\begin{table} [ht]
\caption{Human evaluation of Graph-based model and baseline models.}
\centering
\scalebox{0.75}{
\begin{tabular}{l|ccc}
\hline
Models  & Semantic Relatedness & Fluency & Complexity \\ \hline
NQG++ &   2.86     &     3.22    &    3.06    \\ 
PG &     3.03   &      3.31    &      3.02   \\ 
SM-API &    3.06   &      3.21    &      3.20   \\ \hline
PG + GAT &     3.11   &      3.29    &      3.43   \\ 
\textbf{Proposed} &   \textbf{3.20}  &  \textbf{3.34} &  \textbf{3.71}     \\ \hline
\end{tabular}}
\label{tab:human}
\end{table}

\begin{table*}[t]
\centering
\scalebox{0.9}{
\begin{tabular}{l}
    \hline
    \textbf{Document 1}: [Muriel Humphrey Brown] Muriel Fay Buck Humphrey Brown (February 20, 1912 September 20,\\
    1998) was an American politician who served as the Second Lady of the United States and as a U.S. Senator \\
    from Minnesota Married to the 38th Vice President of the United States, \textcolor{red}{Hubert Humphrey}. \\
    \textbf{Document 2}: [Hubert Humphrey] Hubert Horatio Humphrey Jr. (May 27, 1911January 13, 1978) was an \\
    American politician who served as the 38th Vice President of the United States from 1965 to 1969.\\
    \textbf{Reference}: who is the minnesota senator that was married to muriel humphrey and served as the 38th vice\\
    president of the united states ?\\
    \textbf{PG}: who was an american politician who served as the 38th vice president of the united states from 1965 to 1969 ?\\
    \textbf{SM-API}: who served as the 38th vice president of the united states from 1965 to 1969 ?\\
    \hline
    \textbf{PG+GAT}: who married to Hubert Humphrey who served as the 38th vice president of the united states from \\
    1965 to 1969 ?\\
    \textbf{Proposed}: muriel humphrey brown was an american politician who served as the second lady \\
    of the united states and as a u.s. senator from minnesota married to which american politician who served as\\
    vice president of the united states from 1965 to 1969 ?\\
    \hline
\end{tabular}}
\caption{Case study for showing the superiority of leveraging structured graph data with linguistic relations.}
\label{tab:casestudy}
\end{table*}

As reported in Table \ref{tab:human}, by leveraging the answer-focused multi-hop reasoning, the questions generated by our approach are more complex and semantically relevant to the context and the answer than the baselines.

\paragraph{Case Study}
Table \ref{tab:casestudy} shows question samples generated by the models.
The PG and SM-API fail to discover or capture the entities and their semantic relation from document 1 (e.g., ``Muriel Humphrey married to Hubert Humphrey") and asks the question about the ``Hubert Humphrey served as the 38th vice president of the united states" by only focusing the semantics of the document 2.

However, utilizing the grounded entity graph, GAT-based model generates a more complex question by involving the information of ``Muriel Humphrey married to Huber Humphrey".
Furthermore, by leveraging different granularity levels of the semantic relations among the entities with the answer-focused multi-hop reasoning, the question generated by our model is not only more complex by involving more semantics (e.g., ``Muriel Humphrey married to Huber Humphrey" and ``Muriel Humphrey served as the Second Lady of the United States and as a U.S. Senator from Minnesota.") but also more relevant to the answer than the other models.

\subsection{Implementation Details}
We employ the Spacy toolkit \cite{honnibal2017spacy} to finish the tokenization, NER and POS tagging, and coreference resolution.
We use a 300-dim pre-trained Glove vector as the word embedding.
Following NQG++ \cite{zhou2017neural}, we concatenate the word embedding with 16-dim answer positional features and 16-dim linguistic feature embedding, including the case, NER, and POS tag features.
We train the model of batch size $16$ for $20$ epochs with the Adam optimizer \cite{kingma2014adam} by using a NVIDIA V100 GPU, it takes $80$ minutes to train one epoch.
We use the initial learning rate of $1e-3$ for the model training, and we halve it when the validation BLEU-4 score does not improve on the dev dataset.
We employ beam search with a size of $5$ during the inference.

\section{Related Work}

\paragraph{Single-relation Question Generation}
Existing work dealing with the question generation task can be classified into two categories: rule-based methods and neural network-based methods. 
Rule-based approaches mainly adopt human-designed linguistic templates or rules and are difficult, time-consuming and expensive to scale up. 
Meanwhile, the rigid templates also limit the diversity of the generated questions \cite{mazidi2014linguistic,labutov2015deep}.

Recently, a series of neural network-based models are proposed to solve the problem, as they show a flexible ability of understanding and generating natural language, outperforming the rigid rule-based approaches.
\citet{du2017learning} firstly proposes the question generation task that asks a free question given the context.
Then \citet{zhou2017neural} propose to ask answer-relevant questions given the answer and incorporate the linguistic feature embedding in the model.
\citet{sun2018answer} improve the performance further by utilizing an additional vocabulary for the question word generation and employing the relative answer positional embedding.
\citet{ma2019improving} propose to train two general modules jointly with the seq-to-seq model during training for generating the right keywords and question words and copying the answer-relevant words.

However, the existing models mainly focus on generating questions with the single-relation context.
Different from previous works, we propose a new challenging task to ask complex questions from a collection of documents, requiring the model to discover and reasoning the entities and the semantic relations among them.

\paragraph{Graph Neural Network on NLP tasks}
Leveraging GNN-based models for NLP tasks gained huge popularity recently. 
GNN-based models are mainly adopted to model the semantic and syntactic information from natural language text. 
\citet{zhang2018graph} employs the GCN model to tackle relation extraction on the dependence tree.
A recurrent graph-based model is proposed to solve the bAbI task given the graph-structured input \cite{li2015gated}.
\citet{liu2019learning} apply the GCN model on the dependence tree parsed from the input sentence to predict clue for asking questions. 


\paragraph{Multi-hop Reasoning}
Some works are proposed to realize multi-hop reasoning for question answering given multiple documents. 
\citet{yoon2019propagate} apply Graph Neural Network on a structured graph built with sentences, documents, and query nodes to classify the supporting facts used for answering the query.
\citet{min2019multi} realize the multi-hop reasoning by decomposing the multi-hop query into single-hop queries.
To model the different levels of semantics, a heterogeneous graph consisting of entities, documents, and candidates as nodes \cite{tu2019multi}, which inspires our idea of the answer-centric entity graph.

Different from existing models, our proposed model is always answer-focused during the multi-hop reasoning.
We proposed multi-hop answer-focused reasoning with RGCN facilitates modeling the different levels of semantic information.

\section{Conclusion and Future Work}
In this paper, we proposed a new task that asks complex questions given a collection of documents and the corresponding answer by discovering and modeling the multiple entities and their semantic relations across the documents.
To solve the problem, we propose answer-focused multi-hop reasoning by leveraging different granularity levels of semantic information in the answer-centric entity graph built from natural language text.
Extensive experiment results demonstrate the superiority of our proposed model in terms of automatically computed metrics and human evaluation.
Our work provides a baseline for the new task and sheds light on future work in the multi-hop question generation scenario.

In the future, we would like to investigate whether commonsense knowledge can be incorporated during the multi-hop reasoning to ask reasonable questions.



\bibliography{emnlp2020}
\bibliographystyle{acl_natbib}

\end{document}